\documentclass{article}
\usepackage{spconf,graphicx,hyperref}
\usepackage{amsmath}
\usepackage{amssymb}
\usepackage{mathtools}
\usepackage{amsthm}
\usepackage{bm}
\usepackage{dsfont}
\usepackage{xcolor}
\usepackage{subcaption}

\usepackage{mdframed}
\usepackage{amsthm}
\newtheorem{theorem}{Theorem}
\newtheorem{corollary}{Corollary}
\newtheorem{definition}{Definition}
\newtheorem{assumption}{Assumption}

\usepackage{afterpage}

\usepackage[textsize=tiny]{todonotes}

\usetikzlibrary{arrows,shapes,calc,decorations.pathreplacing,intersections,quotes,angles,positioning,fit,petri,through}
\usepackage{tikz,pgfplots,filecontents}
    \pgfdeclarelayer{background}
    \pgfdeclarelayer{foreground}
    \pgfsetlayers{background,main,foreground}
\usetikzlibrary{spy}
\usetikzlibrary{shapes}
\usetikzlibrary{plotmarks}

\pgfmathdeclarefunction{gauss}{2}{%
   \pgfmathparse{1/(#2*sqrt(2*pi))*exp(-((x-#1)^2)/(2*#2^2))}%
}

\def\R{\mathbb R}

\def\E{\mathbb E}
\def\P{\mathbb P}

\renewcommand\epsilon{\varepsilon}

\newcommand{\tr}{{\rm tr}}

\def\train{{\operatorname{train}}}
\def\test{{\operatorname{test}}}
\def\MAP{{\operatorname{MAP}}}

\def\ve{{\mathbf{e}}}

\def\vv{{\mathbf{v}}}

\def\vx{{\mathbf{x}}}
\def\vy{{\mathbf{y}}}

\renewcommand\epsilon{\varepsilon}

\def\mI{{\mathbf{I}}}

\def\mQ{{\mathbf{Q}}}

\def\mU{{\mathbf{U}}}

\def\mX{{\mathbf{X}}}
\def\mY{{\mathbf{Y}}}

\def\mSigma {{\mathbf{\Sigma }}}
\def\mTheta {{\mathbf{\Theta }}}

\def\R{\mathbb{R}}
\def\P{\mathbb{P}}

\def\E{\mathbb{E}}

\newcommand{\ie}{\emph{i.e.,}~}

\title{Incorporating Priors 
in learning: A Random matrix study under a Teacher--Student Framework}

\name{
  Malik Tiomoko $^{\ast}$, 
  Ekkehard Schnoor $^{\dagger}$
  \thanks{This work was supported by the Research Council of Finland (Decision \#363624) as \textit{A Mathematical Theory of Trustworthy Federated Learning (MATHFUL)}, 
  by the Jane and Aatos Erkko Foundation (Decision \#A835) as
  \textit{A Mathematical Theory of Federated Learning (TRUST-FELT)},
  and by Business Finland as
  \textit{Forward-Looking AI Governance in Banking \& Insurance (FLAIG)}.
  }
}

\address{
    $^{\ast}$ Huawei Noah’s Ark Lab, Huawei Technologies, Paris, France \\
    $^{\dagger}$ Department of Computer Science, Aalto University, Espoo, Finland 
}

\begin{document}
%
\maketitle

\begin{abstract}
Regularized linear regression is central to machine learning, yet its high-dimensional behavior with informative priors remains poorly understood. We provide the first exact asymptotic characterization of training and test risks for maximum a posteriori (MAP) regression with Gaussian priors centered at a domain-informed initialization. Our framework unifies ridge regression, least squares, and prior-informed estimators, and—using random matrix theory—yields closed-form risk formulas that expose the bias–variance–prior tradeoff, explain double descent, and quantify prior mismatch. We also identify a closed-form minimizer of test risk, enabling a simple estimator of the optimal regularization parameter. Simulations confirm the theory with high accuracy. By connecting Bayesian priors, classical regularization, and modern asymptotics, our results provide both conceptual clarity and practical guidance for learning with structured prior knowledge.

\end{abstract}
\begin{keywords}
Random Matrix Theory, Ridge regression, Priors in learning
\end{keywords}

\section{Introduction}
Understanding generalization in high-dimensional regression has become a central theme in modern statistics and machine learning. In the proportional regime ($d,n \to \infty$ with $d/n \to c$), classical intuition breaks down: least squares may overfit catastrophically, ridge regression exhibits double descent, and subtle bias--variance tradeoffs emerge~\cite{bartlett2020benign,hastie2022surprises,belkin2019reconciling}. While recent works have characterized ridge regression and related estimators under Gaussian or sub-Gaussian features~\cite{dicker2016ridge,dobriban2018high,louart2018random}, they largely neglect the use of \emph{informative priors}, despite their prevalence in practice (e.g., pretrained models, temporal smoothness, sparsity) \cite{fortuin2022priors,cheng2022rethinking,cui2022informative}.
In this work, we close this gap by analyzing maximum a posteriori (MAP) regression with Gaussian priors centered at a prior guess $\mTheta_0$. This setting naturally interpolates between ridge regression, the least squares, and prior-informed regression with arbitrary prior. 
Despite its simplicity, the effect of priors on generalization in the high-dimensional limit has remained elusive so far.
Our results can be summarized as follows:
\begin{enumerate}
\itemsep-0.5em 
    \item \textbf{Closed-form asymptotics.} We derive exact formulas for training and test risks under proportional asymptotics, capturing the interaction between prior mismatch, noise variance, and regularization.
    \item \textbf{Unified tradeoff.} The analysis reveals how priors reshape the bias--variance tradeoff, explains double descent, and quantifies the risk as a function of the regularization parameter and the prior quality.
    \item \textbf{Optimal regularization.} We identify the asymptotically optimal regularization parameter in closed form, 
    and propose simple estimators for the test risk, yielding a data-driven rule for near-optimal generalization.
\end{enumerate}

\noindent
Experiments validate the theory, in excellent agreement with the asymptotic predictions. To the best of our knowledge, we provide the first comprehensive analysis of high-dimensional MAP regression with informative priors, extending the frontier of generalization theory and offering practical guidance for leveraging domain knowledge in high dimensions.

\section{ASSUMPTIONS AND SETUP}
\label{sec:setup_and_assumptions}

Let $\{(\vx_i, \vy_i)\}_{i=1}^n \subset \R^d \times \R^q$ be i.i.d. samples generated by
\begin{equation}
\label{eq:teacher_model}
\vy_i = \mTheta_\star^\top \vx_i + \boldsymbol{\varepsilon}_i, \quad 
\boldsymbol{\varepsilon}_i \sim \mathcal{N}(\bm{0}, \sigma^2 \mI_q),
\end{equation}
where $\mTheta_\star \in \R^{d \times q}$ is an unknown weight matrix.  
Denote the design and response matrices collecting the $\vx_i$, $\vy_i$ row-wise,
\[
\mX = \left[ \vx_1^\top; \dots; \vx_n^\top \right] \in \R^{d \times n}, 
\quad
\mY = \left[ \vy_1^\top; \dots; \vy_n^\top \right] \in \R^{q \times n}.
\]
\noindent
We assume access to a prior guess $\mTheta_0 \in \R^{d \times q}$ encoding domain knowledge, like sparsity or temporal patterns.
We obtain the Gaussian likelihood $p(\mY \mid \mX, \mTheta )$, and consider a Gaussian prior $p(\mTheta)$ with $\tilde{\sigma}^2 > 0$ and centered at $\mTheta_0  \in \R^{d \times q}$,
\begin{align*}
p(\mY \mid \mX, \mTheta ) &\sim \prod_{i=1}^n \mathcal{N}(\vy_i \mid  \mTheta^\top \vx_i, \tilde{\sigma}^2 \mI_q), \\
p(\mTheta) &\propto \exp\left(-\frac{\lambda}{2\tilde{\sigma}^2}\|\mTheta-\mTheta_0\|_F^2\right),
\end{align*}
where $p(\mTheta)$ is normalized to be a density function describing a probability distribution over $\mTheta$.
Combining the likelihood and the prior via Bayes' rule next yields the MAP estimator
\begin{align*}
\mTheta_\MAP
&= \arg\min_{\mTheta} 
\frac 1n \|\mY-\mTheta^\top\mX\|_F^2 
+ \lambda \|\mTheta-\mTheta_0\|_F^2 \nonumber \\
&= \left(\frac 1n\mX \mX^\top + \lambda \mI_d \right)^{-1} 
   \left(\frac 1n \mX \mY^\top + \lambda \mTheta_0 \right),
\end{align*}

\noindent
This generalizes classical regression: $\mTheta_0 = \bm{0}$ recovers ridge regression, $\lambda \to 0$ recovers ordinary least squares.
Our main goal is for this generic framework to determine the asymptotically precise training and test risk, defined as follows.

\begin{definition}[Training and Out-of-Sample Risk]
\label{def:training_test_risk}
We define the test risk $\mathcal{R}_\test $ and the training risk $\mathcal{R}_\train$ of $\mTheta_\MAP$ as
\begin{align}
\mathcal{R}_\test &:= \frac{1}{q} \, \E   \left \| \mTheta_\MAP^\top \vx - \vy \right \|_2^2, \label{eq:R_test} \\
\mathcal{R}_\train &:= \frac{1}{qn} \, \E \left \| \mTheta_\MAP^\top  \mX - \mY \right \|_F^2  \label{eq:R_train}
\end{align}
where the expectations are with respect to the training set $\vx_i$ collected in $\mX$, the output noise $\boldsymbol{\varepsilon}_i$, $i =, 1 \dots, n$,
and, for $\mathcal{R}_\test$, additionally with respect to the unseen test input $(\vx,\vy) \sim \mathcal{D}$ sampled from the underlying
distribution $\mathcal{D}$.
\end{definition}

\noindent 
We adopt the following concentration of measure framework.

\begin{assumption}[Data Concentration]
\label{assum:concentration}
The feature vector $\vx_i \in \R^d$ is \emph{$q$-exponentially concentrated}, i.e., for any $1$-Lipschitz continuous (w.r.t. the $\ell_2$-norm) function 
$\varphi : \R^d \to \R$ it holds
\[
\P\left(|\varphi(\vx_i) - \E[\varphi(\vx_i)]| \ge t\right)
\le C e^{-(t/\sigma)^q} \qquad \forall t>0,
\]
for some constants $q>0$, $C>0$, $\sigma>0$ independent of $d$. 
\end{assumption}
\noindent
Random vectors satisfying Assumption \ref{assum:concentration} include isotropic Gaussians, the uniform distribution on the sphere, and any Lipschitz transform thereof (e.g., GAN features \cite{seddik2020random}). Intuitively, any Lipschitz observation of $\vx_i$ is tightly concentrated around its mean, with fluctuations of order $O(1)$ in high dimensions.

\begin{assumption}[Asymptotics]
\label{assum:asymptotic}
As $d, n \to \infty$, the dimension and sample size grow commensurably: $d/n \to c \in (0,\infty)$.  
\end{assumption}

\noindent
W.l.o.g. we assume the data distribution is centered (\ie having zero mean), with the positive definite covariance given by
\begin{equation}
\mSigma = \E \left[ \vx_i\vx_i^\top \right] \in \R^{d \times d}.
\label{eq:Sigma}
\end{equation}

\noindent
This standard setting in modern statistics captures high-dimensional effects (e.g., the curse of dimensionality), while enabling closed-form performance analysis via tools from statistical physics \cite{abarbanel2022statistical} and RMT \cite{cherkaoui2025high,ilbert2024analysing,tiomoko2023pca,tiomoko2022deciphering, couillet2022random}.

\section{Main Results}
\label{sec:main_results}

\noindent The following theorem characterizes the asymptotic test and training risks. Note that they assume knowledge of both the ground-truth teacher $\mTheta_\star$ and the noise variance $\sigma^2$, which are unknown in practice; we adress their estimation in Section~\ref{sec:identity}.

\begin{theorem}[Asymptotic Risks]\label{thm:main_theorem}
Let $(\mu_i, \vv_i), i=1, \dots, d$, be the eigenpairs of the covariance $\mSigma \in \R^{d \times d}$ from \eqref{eq:Sigma}, and  
\begin{align*}
s_i     &:= \left \| (\mTheta_\star - \mTheta_0)^\top \vv_i \right \|_2^2, \\ 
\alpha  &:= \frac{1}{n} \sum_{i=1}^d \frac{\mu_i^2}{(\mu_i + \lambda(1+\delta))^2}, \\
\delta  &=  \frac{1+\delta}{n} \sum_{i=1}^d \frac{\mu_i}{\mu_i + \lambda(1+\delta)} =: f(\delta),
\end{align*}
with an implicit definition of $\delta$ by the fixed-point equation $f(\delta) = \delta$.
Then, under Assumptions \ref{assum:concentration} and \ref{assum:asymptotic}, 
the following risk expressions hold, using the abbreviation $\kappa= \lambda(1+\delta)$.
\end{theorem}
\begin{align*}
\mathcal{R}_\train
&\rightarrow \frac{1+\delta}{q} \left[
\lambda^2 \sum_{i=1}^d \frac{s_i}{\mu_i+\kappa}
\;-\;
\frac{\lambda^3(1+\delta)}{1-\alpha}
\sum_{i=1}^d \frac{s_i}{(\mu_i+\kappa)^2}
\right] \\
&\quad + \sigma^2 \left[
1 - \frac{1}{n}\sum_{i=1}^d \frac{\mu_i}{\mu_i+\kappa}
\;-\;
\frac{\lambda}{n(1-\alpha)} \sum_{i=1}^d \frac{\mu_i}{(\mu_i+\kappa)^2}
\right].
\end{align*}

\begin{align*}
\mathcal{R}_\test 
&\rightarrow \frac{\kappa^2}{1 - \alpha} \cdot \frac{1}{q} 
\sum_{i=1}^d \frac{\mu_i \, s_i}{(\mu_i + \kappa)^2}  \\
& + \sigma^2 \Bigg[
1 + \frac{1+\delta}{n} \sum_{i=1}^d \frac{\mu_i}{\mu_i + \kappa} 
- \frac{\lambda (1+\delta)^2}{(1 - \alpha) n} 
\sum_{i=1}^d \frac{\mu_i}{(\mu_i + \kappa)^2} \Bigg].
\end{align*}

\begin{proof}
We sketch the proof.
The uniqueness of $\delta$ follows from \cite[Proposition 3.12]{louart2018concentration}).
Rearranging \eqref{eq:R_test}, using a basic trace property and taking the expectation with respect to $\vx$,
\[
\mathcal{R}_\test = \frac{1}{q} \E \; \tr\left[  (\mTheta_\MAP- \mTheta_\star) \mSigma  (\mTheta_\MAP- \mTheta_\star)^\top \right] + \sigma^2
\]
we substitute 
$\mTheta_\MAP= \mQ\left(\tfrac1n\, \mX\mY^\top + \lambda \mTheta_0\right)$, 
where the matrix $\mQ = \left(\frac 1n \mX\mX^\top + \lambda \mI\right)^{-1}$ is the so-called resolvent of the matrix $\frac 1n \mX\mX^\top$.
Expand the traces using cyclicity and using the definition of $\mY$ as in \eqref{eq:teacher_model}, and take expectations over the Gaussian noise contained in $\mY$. 
This leaves only terms in $\mSigma$, $\mQ$, and $\mQ\mSigma \mQ$. 
By standard random matrix arguments \cite{louart2018concentration}, we replace $\mQ$ and $\mQ\mSigma \mQ$ by their deterministic equivalents \cite[Definition 4]{couillet2022random}, given as follows by (see also \cite{tiomoko2022deciphering, tiomoko2021advanced, louart2018concentration})
\[
\bar \mQ \leftrightarrow \left(\tfrac{\mSigma}{1+\delta} + \lambda \mI \right)^{-1}, 
\; 
\mQ\mSigma\mQ \leftrightarrow \tfrac{(1+\delta)^2}{(1+\delta)^2 - \tfrac{1}{n} \tr(\mSigma \bar \mQ \mSigma \bar \mQ)} \,\bar \mQ \mSigma \bar \mQ,
\]
with $\delta = \tfrac{1}{n} \tr(\mSigma \bar \mQ)$. 
Diagonalizing $\mSigma = \mU \,\mathrm{diag}(\mu_i)\mU^\top$ and projecting   
$\tfrac1q \mTheta_\star \mTheta_\star^\top$ onto this basis,
$s_i = \tfrac1q  \ve_i^\top \mU^\top \mTheta_\star \mTheta_\star^\top \mU \ve_i$, where $\ve_i$ is the $i$-th standard basis vector in $\R^d$, all trace terms reduce to separable sums $\sum_i f(\mu_i)s_i$ or $\sum_i g(\mu_i)$. 
Collecting coefficients with $\kappa = \lambda(1+\delta)$ yields the closed forms for 
$\mathcal{R}_\train$ and $\mathcal{R}_\test$ as stated.
\end{proof}

\section{Special case of Identity Covariance}
\label{sec:identity} 

In the special case $\mSigma=\mI_d$, the high-dimensional correction $\delta$ admits a closed form, akin to the Marčenko–Pastur law \cite{marvcenko1967distribution}.

\begin{corollary}
Under the conditions of Theorem \ref{thm:main_theorem}, in the case of $\mSigma=\mI_d$, the asymptotic training and test risks are 
\begin{align}
R_\train &\to \frac{1+\delta}{q}\left(\frac{\lambda^2 S}{1+\kappa}-\frac{\lambda^3(1+\delta) S}{(1-\alpha)(1+\kappa)^2}\right) \label{eq:R_train_identity}\\
&\quad + \sigma^2 \left(1 - \frac{c}{1+\kappa} - \frac{\lambda c(1+\delta)}{(1-\alpha)(1+\kappa)^2}\right), \nonumber\\
R_\test &\to \frac{\kappa^2 S}{q(1-\alpha)(1+\kappa)^2} \label{eq:R_test_identity}\\
&\quad + \sigma^2 \left(1 + \frac{c(1+\delta)}{1+\kappa} - \frac{\lambda c(1+\delta)^2}{(1-\alpha)(1+\kappa)^2}\right), \nonumber
\end{align}
where the prior mismatch $S$, and $\delta, \kappa, \alpha$, are given by
\begin{align}
S & = \|\mTheta^\star-\mTheta_0\|_F^2 = \sum_{i=1}^d \left \| (\mTheta_\star - \mTheta_0)^\top \ve_i \right \|_2^2, \label{eq:S}  \\
\delta &= \frac{-(1+\lambda-c) + \sqrt{(1+\lambda-c)^2 + 4\lambda c}}{2\lambda}, \nonumber \\
\kappa &= \lambda(1+\delta), \qquad \alpha = c/(1+\kappa)^2. \nonumber 
\end{align}
\end{corollary}

\begin{figure}[h]
    \centering
    \includegraphics[width=0.4\textwidth]{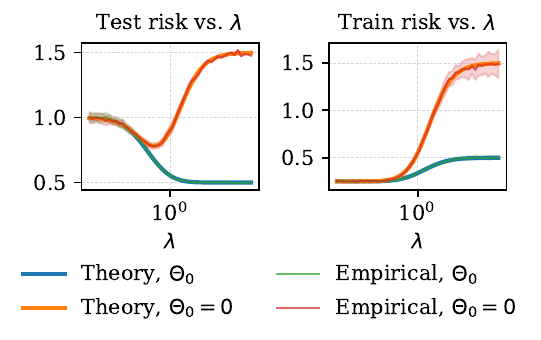}
    \caption{Test (left) and training (right) risks over $\lambda$. Input $d=100$, output $q=10$, training $n=200$, test $N_\text{test}=10\,000$, $\sigma^2=0.5$, $\mSigma=\mI_d$. Both with and without prior on weights.}
    \label{fig:theory_vs_empirical}
\end{figure}

\noindent
Figure\ref{fig:theory_vs_empirical} \footnote{{Reproducibility} code available at \url{https://github.com/maliktiomoko/Priors_learning}} compares theoretical (solid) and empirical (dashed) risks. Blue/green: \(\Theta_0 \neq 0\), orange/red: \(\Theta_0 = 0\), with shaded areas for std.

\paragraph*{Limiting behavior.}  
The extremal limits of these risks are particularly interesting:
\begin{align}
\label{eq:limiting}
\lim_{\lambda \to 0} R_\train       &   = \sigma^2 (1-c), &
\lim_{\lambda \to 0} R_\test        &   = \sigma^2 (1+c), \\
\lim_{\lambda \to \infty} R_\train  &   = \tfrac{S}{q} + \sigma^2, &
\lim_{\lambda \to \infty} R_\test   &   = \tfrac{S}{q} + \sigma^2.
\end{align}
These limits show that as regularization vanishes, the training error is optimistically low while the test error inflates with dimension, whereas strong regularization equalizes them at the signal-to-noise level.
The slopes at the boundaries are
\begin{equation*}
- \left.\frac{dR_\train}{d\lambda}\right|_{\lambda \to 0} 
=
\left.\frac{dR_\test}{d\lambda}\right|_{\lambda \to 0} 
=
\sigma^2 \begin{cases}
\dfrac{c}{1-c}, & 0 < c < 1,\\[2mm]
1, & c > 1,
\end{cases}     
\end{equation*}


These identities are useful for consistent estimation of $\sigma^2$ and $S$ from empirical risk curves.

\begin{figure}[h!]
    \centering
\includegraphics[width=0.5\textwidth]{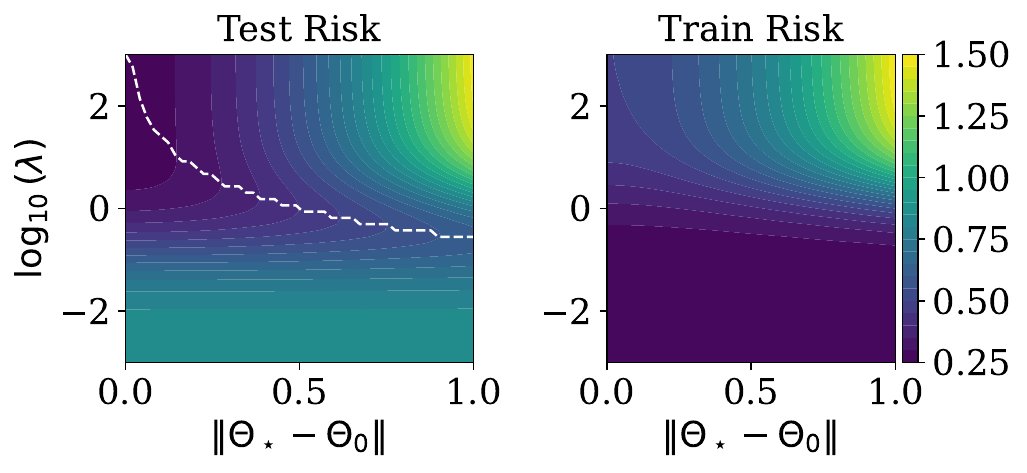}
    \caption{Contour plots of $R_\test$ (left) and $R_\train$ (right) versus prior mismatch $\|\mTheta_\star - \mTheta_0\|$ ($x$-axis) and regularization $\lambda$ ($y$-axis, $\log$-scale); white dashed line: $\lambda$ minimizing $R_\test$; 
    $d=100$, $q=10$, $n=200$, $N_\text{test}=10\,000$,$\sigma^2=0.5$, $\mSigma=\mI_d$.}
    \label{fig:signal_vs_lambda_contour}
\end{figure}

\paragraph*{Prior mismatch and bias-variance tradeoff.} 
We illustrate how the prior mismatch $S$ from \eqref{eq:S} interacts with the regularization parameter $\lambda$ to shape train and test risks. In Figure \ref{fig:signal_vs_lambda_contour}, the plots show the risk landscape, emphasizing regions of high and low risk.

\begin{figure}[h!]
    \centering
    \includegraphics[width=0.5\textwidth]{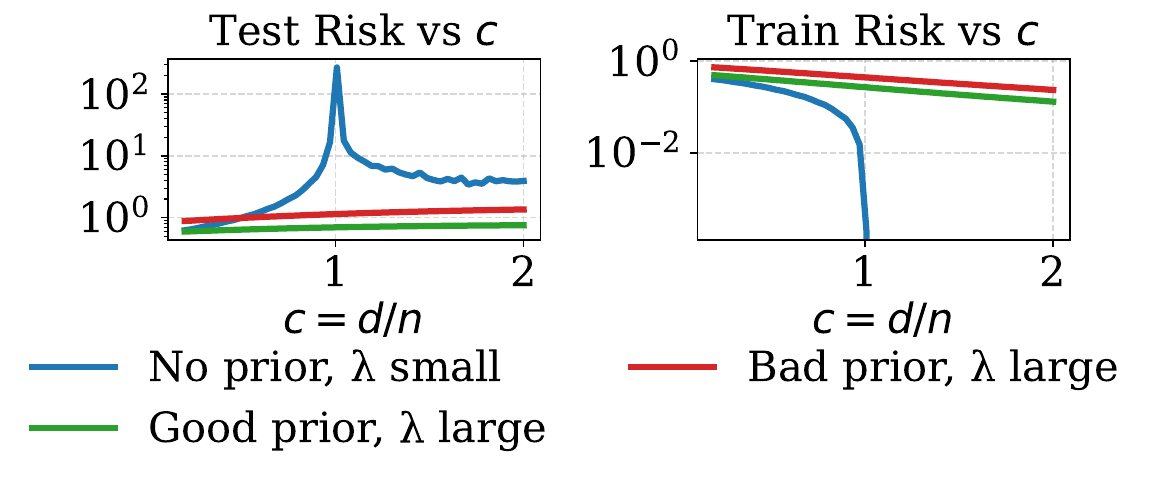} 
    \caption{Test/training risk (left/right) as a function of $c = d/n$ for different priors and regularization strengths; 
    note the double descent phenomenon around the interpolation threshold.}
    \label{fig:double_descent}
\end{figure}

\vspace{-0.5cm}
\paragraph*{Double descent and the role of $1-\alpha$.}  
The denominator $1-\alpha$ in the test risk $R_\test$ from \eqref{eq:R_train_identity}
plays a crucial role in explaining the double descent. When $\lambda\to 0$, $\alpha \to c$, such that $1-\alpha \to 0$ when $c\approx 1$, we obtain a sharp spike in both $R_\train$ and $R_\test$. This explains the interpolation peak in high-dimensional regression: near $c=1$, the variance dominates dramatically. As $\lambda$ increases, $1-\alpha$ grows, and the risks drop, recovering the second descent.
Figure~\ref{fig:double_descent} illustrates the phenomenon of double descent in both test and training risk as a function of the ratio $c = d/n$. The left panel shows the test risk, while the right panel depicts the training risk. In each panel, we consider three scenarios: (i) no prior with a small regularization parameter $\lambda$, (ii) a well-informed prior (\ie small $S$; recall \eqref{eq:S}) with a large $\lambda$, and (iii) a poorly-informed prior (\ie large $S$) with a large $\lambda$. This visualization highlights how the choice of prior and regularization effects the risk landscape, particularly around the interpolation threshold $c \approx 1$, where the test risk exhibits a characteristic peak before decreasing in the overparameterized regime.
\vspace{-0.4cm}
\paragraph*{Optimal regularization.}  
Minimizing $R_\test$ from \eqref{eq:R_test_identity} with respect to $\lambda$ gives the optimal $\lambda^\star$. 
In the underparameterized regime when $c<1$, it assumes the simple closed form
\begin{align}
\lambda^\star = \frac{\sigma^2}{S}\cdot \frac{1}{1-c},
\end{align}
which balances bias and variance explicitly. In the overparameterized regime $c>1$, the optimal $\lambda^\star$ is small but positive, mitigating the variance explosion near the interpolation threshold $c\approx 1$. Intuitively, $\lambda<\lambda^\star$ leads to overfitting, $\lambda>\lambda^\star$ suppresses the signal too much, and $\lambda^\star$ achieves a nearly optimal tradeoff.

\paragraph*{Practical estimation.}  
One can estimate $\sigma^2$ and $S$ from the training curve at extreme $\lambda$ values from Equations \ref{eq:limiting}:
\begin{align*}
\widehat\sigma^2 &\approx \frac{\widehat R_\train(\lambda\to 0)}{1-c}, &
\widehat S &\approx q \left(\widehat R_\train(\lambda\gg 1) - \widehat\sigma^2 \right),
\end{align*}
and then plug these into the formula for $\lambda^\star$ to obtain a data-driven choice that ensures near-optimal generalization.

\begin{figure}[t]
    \centering
\includegraphics[width=0.5\textwidth]{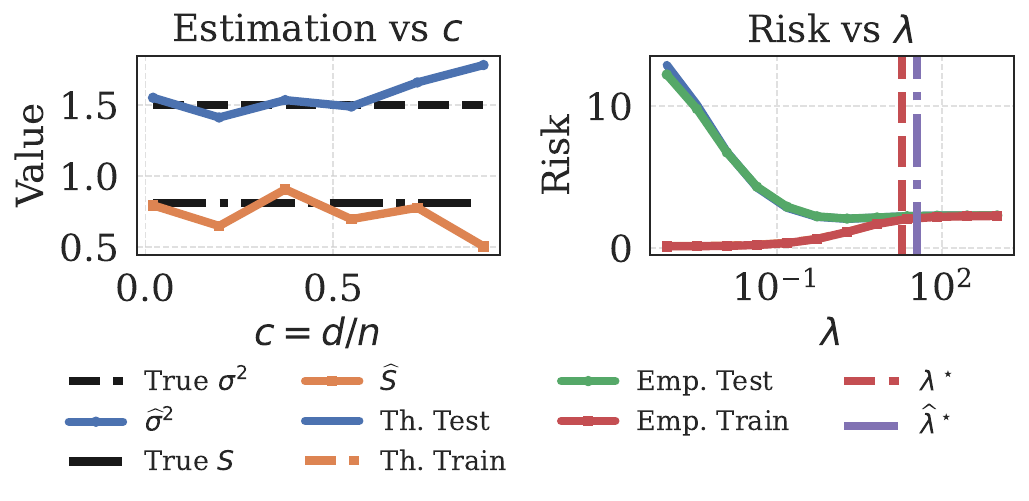}
    \caption{%
        Left: estimation of $\sigma^2$ and $S$ versus $c=d/n$. 
        Right: test risk as a function of $\lambda$, showing both theoretical and empirical curves along with the optimal values of $\lambda^\star$ and $\widehat\lambda^\star$.}
    \label{fig:optimal_lambda}
\end{figure}

Figure~\ref{fig:optimal_lambda} illustrates these theoretical predictions. The left panel shows how $\widehat\sigma^2$ and $\widehat S$ estimated from the training curve track their true values across different aspect ratios $c=d/n$. The right panel compares the theoretical test risk with its empirical counterpart as a function of~$\lambda$, together with the true optimal $\lambda^\star$ and its estimate $\widehat\lambda^\star$. The close agreement confirms that the simple estimators above provide a reliable, data-driven prescription for choosing the regularization parameter.

\section{Extension to A General Covariance}


While in Section \ref{sec:identity} we assumed $\mSigma = \mI$, our main result, Theorem \ref{thm:main_theorem}, is valid for general $\mSigma$.
It is instructive to briefly comment on the differences.
The main subtlety is that the prior mismatch is no longer captured by a single scalar $S$ from \eqref{eq:S},
but instead decomposes along the eigen-directions of $\mSigma$. 
The signal contribution splits into multiple components
\[
s_i = \mathrm{tr}\!\left[(\mTheta^\star-\mTheta_0)\vv_i \vv_i^\top(\mTheta^\star-\mTheta_0)\right],
\]
each weighted by the corresponding $\mu_i$; recall $(\mu_i, \vv_i)$, the eigenpairs of $\mSigma$.
Thus, the signal enters the risk through combinations of several (rather than one)
scalars of the form
\[
\sum_{i=1}^d \frac{\mu_i^{\alpha} s_i^{\beta}}{(\mu_i+\kappa)^{\gamma+1}}, 
\quad (\alpha, \beta,\gamma)\in\{0,1\}^3.
\]
Conceptually, this is the only critical difference: all structural properties of the risks derived in the identity case remain valid. The challenge lies in estimating the signal contribution, which now involves resolving multiple weighted components instead of one. While feasible in principle, obtaining consistent estimators from empirical risk in this setting requires additional work and is beyond the scope of the present paper. We view this as an interesting and promising direction for future research. Another practical distinction is that in the general covariance case the optimal regularization parameter $\lambda^\star$ no longer admits a closed-form expression. Nevertheless, it can be efficiently computed numerically, and the overall qualitative picture of the bias--variance tradeoff remains unchanged.


\section{CONCLUSION AND OUTLOOK}
We provided exact asymptotics for MAP regression with informative priors, unifying ridge, least squares, and prior-informed estimation. Our results clarify the bias--variance--prior tradeoff, explain double descent, and yield practical rules for optimal regularization. Beyond theory, this framework has clear potential for fine-tuning, transfer learning, curriculum learning, and time series forecasting—promising directions for future exploration.

\label{sec:conclusion}

\bibliographystyle{IEEEbib}
\bibliography{strings,refs}

@article{bartlett2020benign,
  title={Benign overfitting in linear regression},
  author={Bartlett, Peter L and Long, Philip M and Lugosi, G{\'a}bor and Tsigler, Alexander},
  journal={Proceedings of the National Academy of Sciences},
  volume={117},
  number={48},
  pages={30063--30070},
  year={2020},
  publisher={National Academy of Sciences}
}

@article{hastie2022surprises,
  title={Surprises in high-dimensional ridgeless least squares interpolation},
  author={Hastie, Trevor and Montanari, Andrea and Rosset, Saharon and Tibshirani, Ryan J},
  journal={Annals of statistics},
  volume={50},
  number={2},
  pages={949},
  year={2022}
}

@article{belkin2019reconciling,
  title={Reconciling modern machine-learning practice and the classical bias--variance trade-off},
  author={Belkin, Mikhail and Hsu, Daniel and Ma, Siyuan and Mandal, Soumik},
  journal={Proceedings of the National Academy of Sciences},
  volume={116},
  number={32},
  pages={15849--15854},
  year={2019},
  publisher={National Academy of Sciences}
}

@article{dicker2016ridge,
  title={Ridge regression and asymptotic minimax estimation over spheres of growing dimension},
  author={Dicker, Lee H},
  year={2016}
}

@article{louart2018random,
  title={A random matrix approach to neural networks},
  author={Louart, Cosme and Liao, Zhenyu and Couillet, Romain},
  journal={The Annals of Applied Probability},
  volume={28},
  number={2},
  pages={1190--1248},
  year={2018},
  publisher={JSTOR}
}

@article{dobriban2018high,
  title={High-dimensional asymptotics of prediction: Ridge regression and classification},
  author={Dobriban, Edgar and Wager, Stefan},
  journal={The Annals of Statistics},
  volume={46},
  number={1},
  pages={247--279},
  year={2018},
  publisher={JSTOR}
}

@inproceedings{tiomoko2023pca,
  title={PCA-based multi-task learning: a random matrix approach},
  author={Tiomoko, Malik and Couillet, Romain and Pascal, Fr{\'e}d{\'e}ric},
  booktitle={International Conference on Machine Learning},
  pages={34280--34300},
  year={2023},
  organization={PMLR}
}

@article{ilbert2024analysing,
  title={Analysing multi-task regression via random matrix theory with application to time series forecasting},
  author={Ilbert, Romain and Tiomoko, Malik and Louart, Cosme and Odonnat, Ambroise and Feofanov, Vasilii and Palpanas, Themis and Redko, Ievgen},
  journal={Advances in Neural Information Processing Systems},
  volume={37},
  pages={115021--115057},
  year={2024}
}

@article{marvcenko1967distribution,
  title={Distribution of eigenvalues for some sets of random matrices},
  author={Mar{\v{c}}enko, Vladimir A and Pastur, Leonid Andreevich},
  journal={Mathematics of the USSR-Sbornik},
  volume={1},
  number={4},
  pages={457},
  year={1967},
  publisher={IOP Publishing}
}

@book{abarbanel2022statistical,
  title={The statistical physics of data assimilation and machine learning},
  author={Abarbanel, Henry DI},
  year={2022},
  publisher={Cambridge University Press}
}

@article{cui2022informative,
  title={Informative Bayesian neural network priors for weak signals},
  author={Cui, Tianyu and Havulinna, Aki and Marttinen, Pekka and Kaski, Samuel},
  journal={Bayesian Analysis},
  volume={17},
  number={4},
  pages={1121--1151},
  year={2022},
  publisher={International Society for Bayesian Analysis}
}

@article{cheng2022rethinking,
  title={Rethinking Bayesian learning for data analysis: The art of prior and inference in sparsity-aware modeling},
  author={Cheng, Lei and Yin, Feng and Theodoridis, Sergios and Chatzis, Sotirios and Chang, Tsung-Hui},
  journal={IEEE Signal Processing Magazine},
  volume={39},
  number={6},
  pages={18--52},
  year={2022},
  publisher={IEEE}
}

@article{fortuin2022priors,
  title={Priors in bayesian deep learning: A review},
  author={Fortuin, Vincent},
  journal={International Statistical Review},
  volume={90},
  number={3},
  pages={563--591},
  year={2022},
  publisher={Wiley Online Library}
}

@article{louart2018concentration,
  title={Concentration of Measure and Large Random Matrices with an application to Sample Covariance Matrices},
  author={Louart, Cosme and Couillet, Romain},
  journal={arXiv preprint arXiv:1805.08295},
  year={2018}
}

@inproceedings{seddik2020random,
  title={Random matrix theory proves that deep learning representations of gan-data behave as gaussian mixtures},
  author={Seddik, Mohamed El Amine and Louart, Cosme and Tamaazousti, Mohamed and Couillet, Romain},
  booktitle={International Conference on Machine Learning},
  pages={8573--8582},
  year={2020},
  organization={PMLR}
}

@inproceedings{tiomoko2022deciphering,
  title={Deciphering lasso-based classification through a large dimensional analysis of the iterative soft-thresholding algorithm},
  author={Tiomoko, Malik and Schnoor, Ekkehard and Seddik, Mohamed El Amine and Colin, Igor and Virmaux, Aladin},
  booktitle={International Conference on Machine Learning},
  pages={21449--21477},
  year={2022},
  organization={PMLR}
}

@phdthesis{tiomoko2021advanced,
  title={Advanced Random Matrix Methods for Machine Learning},
  author={Tiomoko, Malik},
  year={2021},
  school={Universit{\'e} Paris-Saclay}
}

@article{cherkaoui2025high,
  title={High-Dimensional Analysis of Bootstrap Ensemble Classifiers},
  author={Cherkaoui, Hamza and Tiomoko, Malik and Seddik, Mohamed El Amine and Louart, Cosme and Schnoor, Ekkehard and Kegl, Balazs},
  journal={arXiv preprint arXiv:2505.14587},
  year={2025}
}

@book{couillet2022random,
  title={Random matrix methods for machine learning},
  author={Couillet, Romain and Liao, Zhenyu},
  year={2022},
  publisher={Cambridge University Press}
}

\end{document}